\documentclass[runningheads]{llncs}
\usepackage[T1]{fontenc}
\usepackage{graphicx}
\usepackage{booktabs}
\usepackage[misc]{ifsym}
\newcommand{\corr}{(\Letter)}
\usepackage{mwe}
\usepackage{multirow}
\usepackage{placeins}
\usepackage[normalem]{ulem}
\usepackage{color}
\usepackage{hyperref}
\usepackage{xurl}

\begin{document}
	
	\title{A High-Resolution Landscape Dataset for Concept-Based XAI With Application to Species Distribution Models}

	\titlerunning{A High-Resolution Landscape Dataset for Concept-Based XAI}	
	\author{Augustin de la Brosse\inst{1, 2} \corr \and
		Damien Garreau\inst{3} \and
		Thomas Houet\inst{1, 2} \and
		Thomas Corpetti\inst{1}}
	
	\authorrunning{A. de la Brosse et al.}
	
	\institute{Université Rennes 2, CNRS, Nantes Université, Univ Brest, LETG, UMR 6554, 35000 Rennes, France \email{$\{$augustin.de-la-brosse, thomas.houet,thomas.corpetti$\}$@univ-rennes2.edu}
		\and
		LTSER Zone Atelier Armorique, Rennes, France
		\and
		University of Würzburg, Center for Artificial Intelligence and Data Science, Germany
		\email{$\{$damien.garreau$\}$@uni-wuerzburg.de}}
		
	\tocauthor{de la Brosse,Garreau,Houet,Corpetti}
	\toctitle{A High-Resolution Landscape Dataset for Concept-Based XAI With Application to Species Distribution Models}
	
	\maketitle

	\begin{abstract}
		
		Mapping the spatial distribution of species is essential for conservation policy and invasive species management. Species distribution models (SDMs) are the primary tools for this task, serving two purposes: achieving robust predictive performance while providing ecological insights into the driving factors of distribution. However, the increasing complexity of deep learning SDMs has made extracting these insights more challenging. To reconcile these objectives, we propose the first implementation of concept-based Explainable AI (XAI) for SDMs. We leverage the Robust TCAV (Testing with Concept Activation Vectors) methodology to quantify the influence of landscape concepts on model predictions. To enable this, we provide a new open-access landscape concept dataset derived from high-resolution multispectral and LiDAR drone imagery. It includes 653 patches across 15 distinct landscape concepts and 1,450 random reference patches, designed to suit a wide range of species. We demonstrate this approach through a case study of two aquatic insects, Plecoptera and Trichoptera, using two Convolutional Neural Networks and one Vision Transformer. Results show that concept-based XAI helps validate SDMs against expert knowledge while uncovering novel associations that generate new ecological hypotheses. Robust TCAV also provides landscape-level information, useful for policy-making and land management. Code and datasets are publicly available.
		
	\end{abstract}

	\section{Introduction}\label{intro}
	
	\begin{figure}[hbt!]
		\includegraphics[width=\textwidth]{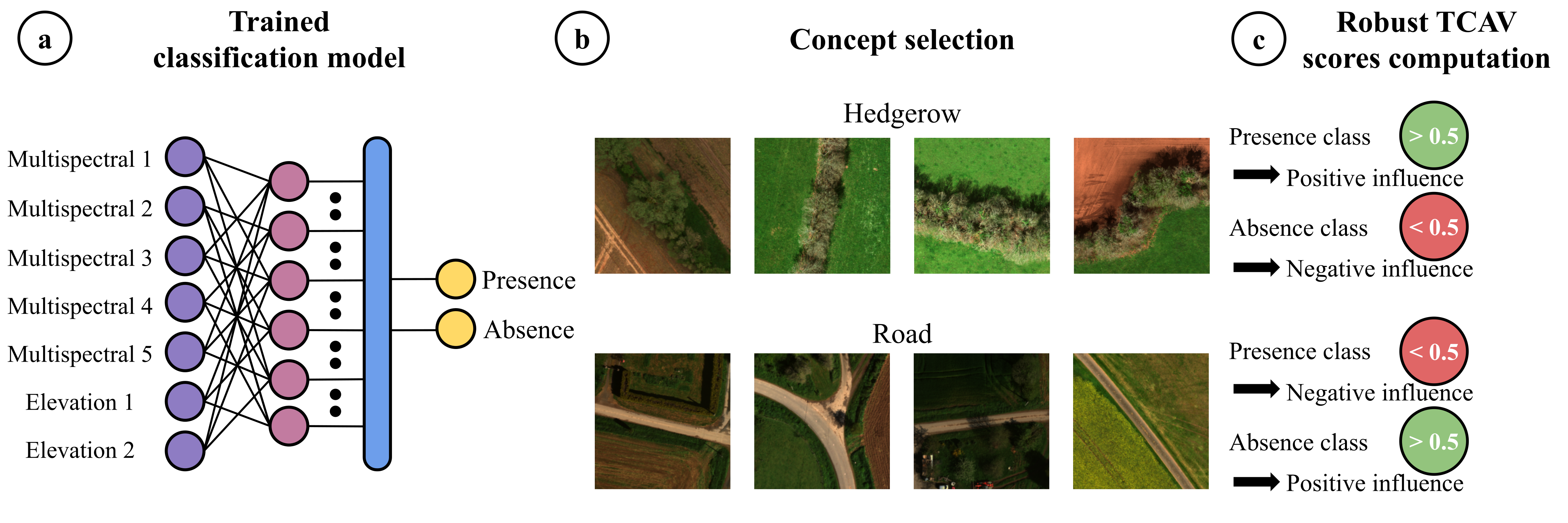}
		\caption{\textbf{Schematic overview of our implementation of Robust TCAV}: After training a model to classify the classes \texttt{Presence} and \texttt{Absence} of insect \textcircled{a}, we select concepts (\emph{e.g.}, hedgerow) \textcircled{b}, and quantify its importance for the chosen class by computing its Robust TCAV score \textcircled{c}. 
		} \label{fig_intro}
	\end{figure}
	
	Species Distribution Models (SDMs) provide a framework for spatially mapping species. These models integrate field observations with a suite of environmental and spatial covariates to predict the presence and absence of the studied species \cite{elith2009species}. In doing so, SDMs serve a dual purpose: achieving robust predictive performance while providing ecological insights into the drivers of distribution. Consequently, SDMs are essential for a variety of applications, ranging from conservation policy planning to risk assessment for invasive species. SDMs often rely on traditional machine learning (ML) models such as Random Forest \cite{gerber2023landscape} to predict species distribution and explainable artificial intelligence (XAI) techniques to extract knowledge from models \cite{islam2022past}. Over the last decade, the adoption of complex deep learning algorithms and high-dimensional remote sensing data has improved predictive performance compared to traditional models, as shown by the GeoLifeCLEF challenge results \cite{botella2025overview,lorieul2022overview,picek2024overview}. However, unlike traditional models where feature importance can be directly assessed, deep learning models learn abstract features across millions of parameters, making it increasingly challenging for SDM developers to achieve the objective of knowledge extraction.  
	
	Fortunately, the artificial intelligence (AI) community has developed different XAI methods that can be used to interpret deep SDMs \cite{islam2022past}. Among them, concept-based XAI methods \cite{poeta2023concept} aim to align with human reasoning by explaining AI predictions through high-level, interpretable concepts (\emph{e.g.}, colors). This facilitates interdisciplinary collaborations, as members of other scientific communities may not be ML experts, but may contribute to defining the concepts and easily understand the output. However, implementing these methods requires a dedicated concept dataset, which currently does not exist for SDMs.
	
	In this paper, we aim to bridge the gap between predictive performance and  interpretability in deep SDMs. We leverage drone data to create a unique concept dataset adapted to fine-scale SDMs. Drones have become increasingly used in ecological studies due to their flexibility in providing very-high spatial, spectral, and temporal resolution data \cite{anderson2013lightweight}. By using multispectral and LiDAR-derived drone imagery, we extract landscape concepts that are relevant to explaining multiple ecological processes.
	
	We apply concept-based XAI to SDMs using two taxa, Plecoptera and Trichoptera, as a case study. They are emerging aquatic insects: after a larval stage underwater, they emerge and disperse around their water body of origin. During this dispersal, they provide a number of ecosystem services, \emph{i.e.}, the direct and indirect contributions of ecosystems to human welfare \cite{raitif2019stream}.
	
	To contribute to the reconciliation of the two original objectives (interpretation and accurate predictions) of SDMs, we make the following contributions:
	\begin{itemize}    
		\item We release a \textbf{new high-resolution concept dataset}. It includes 15 landscape concepts (653 patches in total) and 1,450 random patches, derived from multispectral and LiDAR drone data;
		\item We propose \textbf{three new SDMs for two aquatic insects}, Plecoptera and Trichoptera. These SDMs are based on distinct architectures (two convolutional neural networks and a vision transformer) adapted to leverage high-resolution multispectral and LiDAR drone imagery;
		\item We show that Robust TCAV \cite{martin2019interpretable}, combined with our new dataset, provides ecologically meaningful explanations from these SDMs, constituting the \textbf{first implementation of concept-based XAI to SDMs} (Fig. \ref{fig_intro}).
	\end{itemize}
	
	The paper is structured as follows: Section \ref{dataset} describes our concept (see Fig.~\ref{dataset_patch}) and training datasets. Section \ref{experiments} details the experimental setup and presents the results for both the SDMs and the Robust TCAV implementation. Finally, Section \ref{discussion} discusses key findings, and Section \ref{conclusion} concludes the study.
	
	\subsection{Related work}
	Attribution methods such as Local Interpretable Model-agnostic Explanations (LIME) \cite{ribeiro2016should} and Shapley values (SHAP) \cite{lundberg2017unified} have been extensively applied to traditional SDMs \cite{buebos2025evaluating,he2022explainable,ryo2021explainable} to extract insights from tabular features. However, when applied to image-based models, these methods operate at the pixel level, highlighting which individual pixels contributed most to a single prediction. This makes them both challenging to interpret for non-expert users and hinders the ability to provide the global, higher-level explanations that ecologists need to derive actionable insights \cite{poeta2023concept}. Notably, they have not been used to explain deep SDMs. To date, MaskSDM \cite{zbinden2026masksdm}, which is explainable by design, is the only work trying to reunite high performance with interpretability in the SDM domain.
	
	As mentioned earlier, concept-based XAI represents a promising solution, with Testing with concept activation vectors (TCAV) \cite{pmlr-v80-kim18d} as a seminal work. This post-hoc approach relies on a Concept Activation Vector (CAV) which represents the direction of the concept in the latent space of the model. The TCAV score quantifies how much a prediction aligns with this concept direction. This method aims to quantify the influence of concepts in a classification result. Since its introduction, several improvements, such as Robust TCAV \cite{martin2019interpretable}, Adversarial-TCAV \cite{soni2020adversarial}, Concept Activation Regions \cite{crabbe2022concept}, or pattern-based CAVs \cite{ICLR2025_852870d5}, have been proposed to increase its robustness. To our knowledge, only three ecological studies implement TCAV or its adaptations \cite{amara2024explainability,amara2023concept,huang2025barkxai}. While none apply it to SDMs, all three indicate that the results align with expert knowledge.

	\section{A high-resolution dataset for concept-based XAI}\label{dataset}
	
	In this subsection, we first describe the creation of a new and public concept dataset\footnote{Access to dataset: \url{https://zenodo.org/records/18936778}.} to help SDM practitioners gain ecological insights from their models. Secondly, we describe the species distribution data.

	\subsection{Concept dataset ($C_i$)}\label{concept_dataset}
	
	\textbf{Concept selection.}
	We drew on the multidisciplinary expertise of our team to select 15 concepts inspired from land use / land cover classes. These concept are commonly used to understand ecological processes. We outline here these concepts and indicate how they are relevant to studying the distribution of Plecoptera and Trichoptera, as well as a wide range of species.
	\begin{itemize}
		\item \textbf{Hedgerows} (\texttt{Hedge}) are linear forests composed of different layers of vegetation. They contribute to the proper functioning of ecosystems by protecting crops, contributing to biological diversity and soil conservation \cite{montgomery2020hedgerows};
		\item \textbf{Isolated trees} (\texttt{IsoTree}) are a characteristic feature of the agricultural landscape, an island of biodiversity within open and cultivated environments~\cite{prevedello2018importance}; 
		\item \textbf{Woodlands} (\texttt{Wood}) are an essential landscape component. They are a refuge for biodiversity while also delivering several services such as pest control~\cite{valdes2020high};
		\item \textbf{Linear bodies of water} (\texttt{LinW}) and \textbf{surface bodies of water} (\texttt{SurfW}) shelter significant biodiversity and concentrate many energy fluxes between aquatic and terrestrial ecosystems \cite{bergerot2025tightly}. Linear bodies of water include permanent and temporary bodies: rivers, streams, stagnant and flowing ditches;
		\item \textbf{Wetlands} (\texttt{Wet}) represent ecosystems where water serves as the fundamental driver of environmental conditions and biological composition. These areas provide support for biodiversity and enhance water quality \cite{zedler2005wetland};
		\item \textbf{Roads} (\texttt{Road}) and \textbf{buildings} (\texttt{Build}) represent a major human-induced transformation of landscapes as they contribute to destroying and fragmenting them. Thus, they change the biotic and abiotic conditions of the landscape, often resulting in the modification of species composition~\cite{marcantonio2013biodiversity};
		\item \textbf{Wheat crops} (\texttt{Wheat}), \textbf{maize crops} (\texttt{Maize}), and \textbf{other cereal crops} (\texttt{Cereal}) (millet, sorghum, durum wheat, buckwheat, and winter rye) correspond to a large proportion of agricultural landscapes. Different types of crop may have very different effects \cite{fan2024impact}; 
		\item \textbf{Permanent grasslands} (\texttt{PermG}) and \textbf{temporary grasslands} (\texttt{TempG}) are grassy expanses in a temperate climate. They have the potential to be among the most species-rich habitats globally while also delivering important ecosystem services such as pollination or pest control \cite{wilson2012plant};
		\item \textbf{Organic crops} (\texttt{Organic}) are defined as a method of agricultural production that excludes chemical inputs, such as fertilizers or pesticides, and genetically modified organisms. As such, they are often characterized by increases in species abundance and richness compared to conventional crops~\cite{hole2005does};
		\item \textbf{Conventional crops} (\texttt{Convent}) are the most common type of crop globally. They rely on chemical input. Therefore, they are often negatively correlated with species richness and abundance \cite{leroy2026acceleration}.
	\end{itemize}
	
	\begin{table}[tb]
		\caption{List of the 15 concepts as well as the number of patches.}
		\label{tab:concept_description}
		\centering
		\begin{tabular}{l@{\hskip 0.5cm}c@{\hskip 0.5cm}c}
			\toprule
			Concept & Name & \shortstack{\# of patches \\ (\% of total number of patches)}  \\
			\midrule
			Hedgerow                & \texttt{Hedge}   & 50 (7.7\%) \\
			Isolated tree           & \texttt{IsoTree} & 45 (6.9\%) \\
			Woodland                & \texttt{Wood}    & 50 (7.7\%)  \\
			Linear body of water    & \texttt{LinW}    & 50 (7.7\%)   \\
			Surface body of water   & \texttt{SurfW}   & 20 (3.1\%) \\
			Wetland                 & \texttt{Wet}     & 50 (7.7\%)  \\
			Road                    & \texttt{Road}    & 49 (7.5\%) \\
			Building                & \texttt{Build}   & 50 (7.7\%)    \\
			Wheat crop              & \texttt{Wheat}   & 50 (7.7\%)   \\
			Maize crop              & \texttt{Maize}   & 50 (7.7\%)   \\
			Other cereal crop       & \texttt{Cereal}  & 20 (3.1\%)   \\
			Permanent grassland     & \texttt{PermG}   & 39 (6.0\%)  \\
			Temporary grassland     & \texttt{TempG}   & 50 (7.7\%)  \\
			Organic crop            & \texttt{Organic} & 30 (4.6\%) \\
			Conventional crop       & \texttt{Convent} & 50 (7.7\%)   \\
			\bottomrule
		\end{tabular}
	\end{table}
	
	\medskip
	\noindent\textbf{Concept patch creation.} We created the concept patches from high-resolution drone data acquired in April 2024 at five study sites in France using a Trinity F90+ drone. It was equipped with a MicaSense Dual MX sensor to acquire 5-band multispectral data, and a Qube240 sensor to acquire LiDAR data. The spectral bands are blue (475 nm $\pm$ 32), green (560 nm $\pm$ 27), red (668 nm $\pm$ 20), red-edge (717 nm $\pm$ 12), and near-infrared (842 nm $\pm$ 57). The spatial resolution is 8 cm/pixel for flight at 120 m above ground level. For each spectral band, orthophotographs (\emph{i.e.}, photographs free from topographic distortion) were generated and radiometrically corrected. This process accounted for solar irradiance and angle using a calibrated radiometric target to provide top-of-canopy reflectance \cite{assmann2018vegetation}. LiDAR data are characterized by a 3D point cloud (average 36.1 points per square meter) from which we derived a digital terrain model (elevation at ground level) and a digital surface model (elevation at canopy level) with a spatial resolution of 10 cm/pixel. LiDAR and multispectral data are registered using common ground control points whose positions are set using a GNSS (2-3 cm accuracy). Geometric errors are smaller than the pixel size.

	\begin{figure}[tb]
		\includegraphics[width=\textwidth]{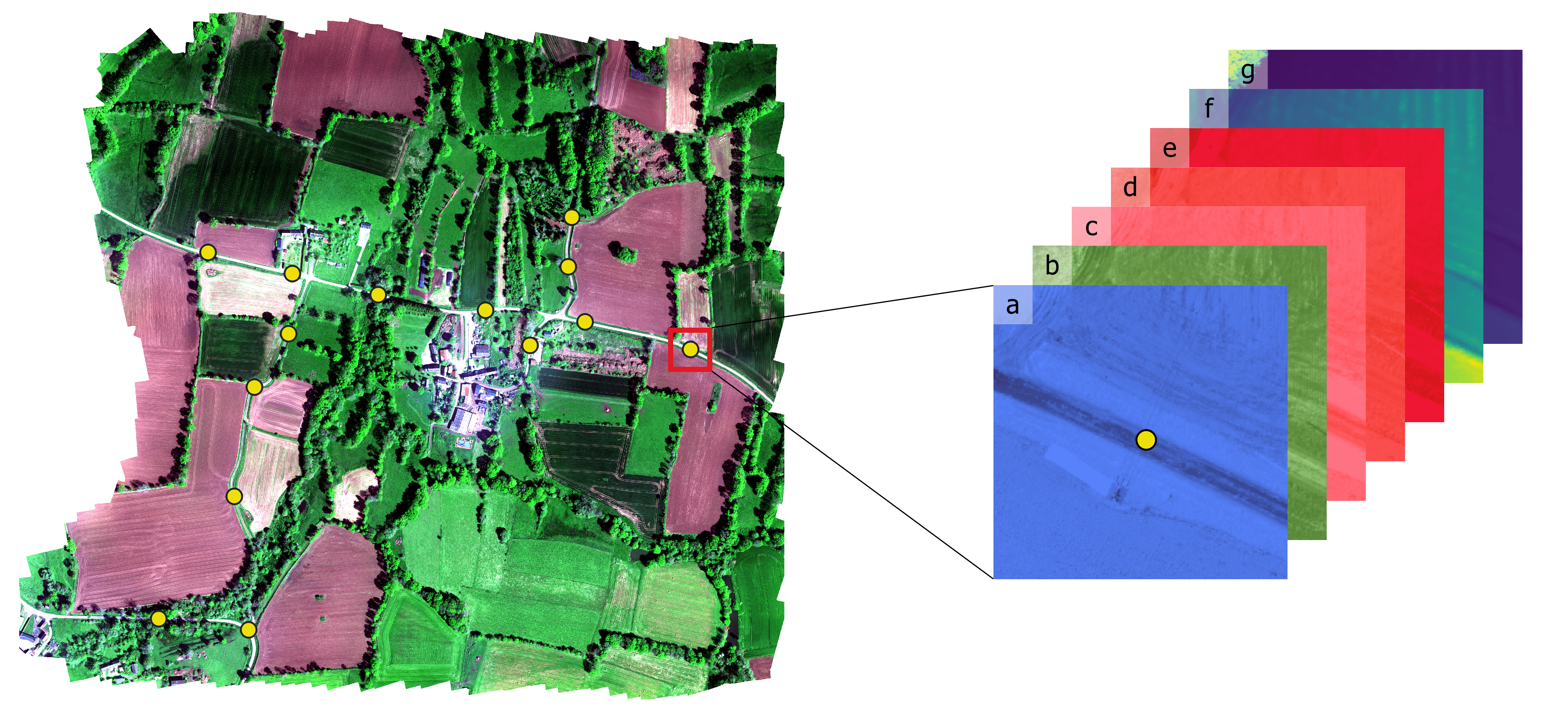}
		\caption{\textit{Left panel}: global overview of one of the acquisition sites. Yellow dots correspond to the locations where \texttt{Road} concept patches were extracted. \textit{Right panel}: The patch is extracted around the point vector from the multispectral image and the digital elevation models to obtain a 7-band patch.
		} \label{dataset_patch}
	\end{figure}

	To create the patches, we first located concepts using point vectors in QGIS\footnote{QGIS is an open source geographic information system (GIS) software that allows to view, edit, print, and analyze geospatial data in various formats.}. After resampling the multispectral data and the elevation models to a common spatial resolution (8 cm/pixel), we extracted from both modalities 512 $\times$ 512 pixel patches centered on the points, stacking five multispectral bands and two elevation models (respectively, bands a-e and bands f-g in Fig.~\ref{dataset_patch}). 
	
	In total, we created 653 concept patches (see Table \ref{tab:concept_description} for the number of patches per concept). By extracting such data from five study sites, we ensured a good representation of a diversity of agricultural landscapes, ranging from extensive dairy farming systems to highly intensive cropping systems. 
	
	Creating the concept dataset was particularly challenging. Indeed, many landscape elements cannot be identified through photointerpretation \cite{lillesand2015remote} of the drone images. In particular, the linear and surface bodies of water are often below the canopy or too small/thin to be observed. The wetlands are also often visually covered by vegetation. Therefore, we relied on the manual mapping of the hydrographic network of the five study sites. To differentiate between the various types of crop and grassland, we used the 2024 French official agricultural land register\footnote{\url{https://geoservices.ign.fr/rpg}}. Regarding the organic and conventional crops, we used the 2024 register of the French agency Agence Bio\footnote{\url{https://www.data.gouv.fr/datasets/parcelles-certifiees-en-agriculture-biologique-sur-cartobio}}. We conducted the identification of the hedgerows, isolated trees and woodlands through photointerpretation. We checked the digital height surface (the discrepancy between the surface and terrain models) to ensure that the vegetation height is over 2.5 meters. Finally, the buildings and the roads were identified by photointerpretation alone.
	
	We also added 1,450 random patches extracted from the multispectral data and the digital elevation models by sampling random pixel coordinates and then applying the same methodology as the one employed for the concept patches.

	To ensure the internal consistency and distinctiveness of our concepts, we performed a series of sanity checks (Section~\ref{concept_sanity_check} of the Appendix). These checks included evaluating the separability of representative concepts against random patches and applying Robust TCAV to confirm that the Concept Activation Vector (CAV) of each concept positively influences its own classification.

	\subsection{Species distribution data}
	
	We collected distribution data for Plecoptera and Trichoptera across the same five study sites used for concept extraction.\footnote{Access to distribution data: \url{https://zenodo.org/records/18937048}.}
	
	\medskip
	\noindent\textbf{Training labels ($Y_i$).} The labels for our two classes are \texttt{Presence} and \texttt{Absence}, for Plecoptera and Trichoptera. This information was obtained in April 2024 from 234 sticky traps uniformly distributed at the study sites. The acquisition period coincides with the primary emergence period for Plecoptera and Trichoptera. \texttt{Presence} was recorded if at least one insect was captured during the two-week trap deployment. Otherwise, \texttt{Absence} was noted. The sample size reflects the inherent logistical constraints of field-based ecological research ~\cite{gerber2023landscape,silva2019current,tognelli2009evaluation}. The trap analysis was conducted following the same protocol as in~\cite{gerber2023landscape}.
	
	\medskip
	\noindent\textbf{Training patches ($X_i$).} We generated training patches from the same drone imagery used for our concept patches. For each sampling point (\emph{i.e.}, sticky trap), we extracted 7-band patches centered on the trap following the extraction protocol previously described (see Section \ref{concept_dataset}). The patch size may be smaller than 512 $\times$ 512 if the trap is closer to the border of the multispectral images and digital elevation models than 256 pixels. These patches served as input for training both the Plecoptera and Trichoptera models.

	\section{Models, experimental setup and results} \label{experiments}
	
	In this section, we evaluate three SDMs with respect to the predictive performance for \texttt{Presence} and \texttt{Absence} of Plecoptera and Trichoptera before implementing Robust TCAV\footnote{Access to code: \url{https://github.com/augustin-delabrosse/Robust-TCAV-for-SDM}.}.
	
	\subsection{Model Architectures}\label{models}
	We selected two CNNs and one Vision Transformer to train the SDMs.
	
	\textbf{CerberusCNN.} This custom architecture has three parallel branches to capture spatial features at multiple scales. Each branch uses two convolutional blocks with distinct kernel sizes (3, 5, and~7) and initial max-pooling sizes (2, 4, and 8) to detect patterns across varying scales. Features are aggregated via 2D global average pooling, concatenated, and processed through a classification head consisting of two dense layers (256 and 128 neurons) with intermediate dropout (0.5 and 0.3). It has approximately 270k learnable parameters.

	\textbf{Adapted ResNet-50 (ARN-50).} As a benchmark for deep hierarchical feature learning, we used ResNet-50 \cite{he2016deep}, an architecture commonly used in ecology \cite{hindarto2023use,kaur2024enhanced,sankupellay2018bird}. The architecture was modified by replacing the initial layer to accommodate 7-channel inputs and appending a two-neuron linear layer for binary classification. It has approximately learnable 23.5M parameters.
	
	\textbf{PicoViT.} In order to assess the efficacy of attention mechanisms in SDMs, we adapted the Vision Transformer (ViT) architecture \cite{dosovitskiy2021an}. We significantly reduced the model capacity to approximately 146k learnable parameters to suit our dataset scale, employing 3 layers with 3 attention heads each. The embedding and hidden MLP dimensions were set to 48 and 96, respectively, with a patch size of 16. Like the ARN-50 model, the input layer was modified to process 7-channel inputs.
	
	An illustration of CerberusCNN is provided in the Appendix (Section~\ref{cerberuscnn_illustration}). Please refer to \cite{he2016deep} and \cite{dosovitskiy2021an} for more details concerning the ARN-50 and the PicoViT, respectively.
	
	\subsection{Training Setup}\label{training_conf}
	Models were trained independently for each taxon using cross-entropy loss and AdamW optimizer \cite{loshchilov2017fixing} (learning rate: 0.001, weight decay: 0.0001, default $\beta_1$ and $\beta_2$). We set batch size to 8 and number of epochs to 50 with a 5-epoch patience with respect to validation loss as early-stopping. For pretrained ARN-50, ImageNet weights \cite{he2016deep} were used for the RGB channels, while the four additional channels were initialized as the mean of the pretrained RGB weights before full-model fine-tuning.
	
	\noindent\textbf{Patch Preprocessing.} To prevent data leakage \cite{guelat2018effects} caused by overlapping of spatially close patches, we implemented a longitudinal spatial split of the drone imagery before extracting patches. We used the easternmost 20\% (48 occurrences) of traps for testing, the next 20\% for validation (38 occurrences), and the remainder for training (198 occurrences). Label distribution remains roughly balanced considering the small size of the dataset (Section~\ref{label_distrib} of the Appendix). Patches were resized via bilinear interpolation to $128 \times 128$ pixels, clipped at zero to remove \texttt{no data} artifacts, and Min-Max normalized per band.
	
	\noindent\textbf{Data Augmentation.} We applied data augmentation to our dataset to increase the number of training samples and improve regularization. We used weak random transformations (horizontal and vertical flips, rotations). We also used MixUp \cite{zhang2018mixup}, which linearly combines pairs of images and their labels to generate new synthetic samples. 
	
	\subsection{Predictive Performance}
	
    \begin{table}[tb]
    	\caption{Model predictive performance per taxon: mean AUC and standard deviation over three training runs. Best scores are in bold.}
    	\label{classif_results}
    	\centering
    	\begin{tabular}{lcccc}
    		\toprule
    		Taxon  & CerberusCNN & ARN-50 & Pretrained ARN-50 & PicoViT \\
    		\midrule
    		Plecoptera    & 0.88 (0.01)  & 0.82 (0.04)  & \textbf{0.90} (0.07)  & 0.72 (0.02) \\
    		Trichoptera   & \textbf{0.79} (0.03)  & 0.74 (0.04)  & 0.75 (0.08)  & 0.66 (0.06) \\
    		\bottomrule
    	\end{tabular}
    \end{table}
    
	We use the Area Under Curve (AUC) as the evaluation metric as it is frequently used by ecologists \cite{gerber2023landscape}. In ecology, a model is considered reliable when its AUC is over 0.7 and excellent when it is over 0.8 \cite{guisan2017habitat}.
	
	Table~\ref{classif_results} shows that for all models, Plecoptera is better predicted than Trichoptera. Among models trained from scratch, CerberusCNN achieves the best results, yielding reliable AUCs for both taxa. ARN-50 also provides reliable predictions (0.82 and 0.74 AUC for Plecoptera and Trichoptera, respectively). PicoViT obtains a reliable AUC for Plecoptera (0.72 AUC), but remains unreliable for Trichoptera (0.66 AUC). When pretrained, ARN-50 outperforms all other models for Plecoptera (0.90 AUC) and improves the AUC for Trichoptera (0.75 AUC). Overall, CerberusCNN has the lowest standard deviation for both taxa. We include distribution maps in the Appendix (Section~\ref{classification_maps}).

	\subsection{Robust TCAV Configuration}
	
	\begin{figure}[htb!]
		\includegraphics[width=\textwidth]{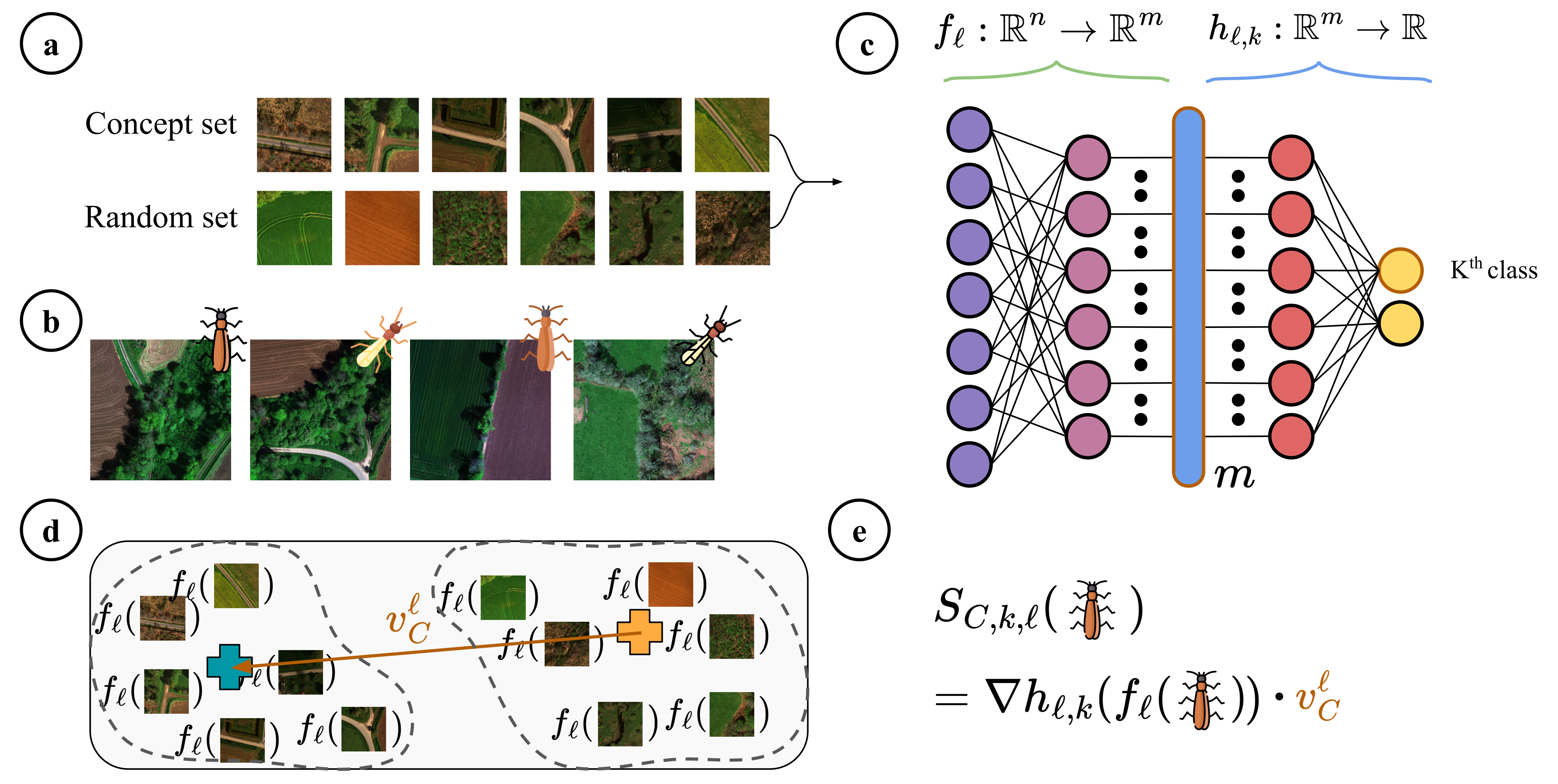}
		\caption{\textbf{Robust TCAV:} Robust TCAV requires a set of examples for a concept (\emph{e.g.}, \texttt{Road}), and random examples \textcircled{a}, labeled test examples for the studied class (.\emph{e.g.}, \texttt{Presence} of Plecoptera) \textcircled{b} and a trained network \textcircled{c}. We obtain the CAV ($v^{\ell}_C$, orange arrow) by computing the difference between the mean activations for the concept examples and for the random examples \textcircled{d}. We use the directional derivative to quantify the sensitivity score $S_{C,k,\ell}$ of an individual training example to concept $C$ for the class $k$. Robust TCAV score for class $k$ is the proportion of positive sensitivity scores among input examples of this class \textcircled{e}.} \label{fig_tcav}
	\end{figure}
	
	To quantify concept influence for each taxon, we compute TCAV scores using the Robust TCAV methodology \cite{martin2019interpretable} on CerberusCNN, Pretrained ARN-50, and PicoViT (Fig.~\ref{fig_tcav}). While TCAV \cite{pmlr-v80-kim18d} defines the concept activation vector (CAV) as the normal vector to the hyperplane of the linear classifier that separates random patches from concept ones, Robust TCAV instead uses the difference between mean concept and random activations. This improves the consistency of the TCAV scores and reduces sensitivity to noise in the concept datasets \cite{martin2019interpretable}.

	\medskip
	\noindent\textbf{Robust TCAV.} Several parameters need to be specified. First, for each model, we selected a layer $\ell$ to extract the activation values and the gradient values. For CerberusCNN, we selected the concatenation of the three branches as it gathers information from all branches. For ARN-50, we used the last bottleneck block before the average pooling layer as deeper layers of CNNs process higher-level features than early layers \cite{ilin2017abstraction,zeiler2014visualizing,zhong2019shallow}. For PicoViT, we used the middle layer (2nd layer) as the middle layers of Transformers allow for the extraction of high-level feature representations \cite{ICLR2025_57568e09,NEURIPS2023_a0e66093,zou2023representation}. Secondly, we computed TCAV scores 500 times as proposed by \cite{pmlr-v80-kim18d}. For each iteration, we sampled 500 random patches. Finally, while Robust TCAV applies a 0.05 threshold on the sensitivity scores (see Fig.~\ref{fig_tcav}), we found that a zero-threshold provides more ecologically relevant insights. Conversely, applying a threshold discards low-magnitude but biologically significant signals, which results in ambiguous results. More details about each step and the threshold are available in the Appendix  (Section~\ref{threshold}).
	
	\medskip
	\noindent\textbf{Relative importance.} Relative TCAV was introduced by \cite{pmlr-v80-kim18d}. It allows the pairwise comparison of concepts. In this scenario, a linear classifier is trained to separate two concepts and the normal vector to the hyperplane is used as CAV. Although \cite{martin2019interpretable} do not adapt this module for Robust TCAV, we replaced the normal vector with the difference of means for consistency. In this setting, the CAV no longer represents the direction of a concept relative to a random set, but rather the direction of one concept relative to another.  
	To obtain a concept ranking, we counted how many times each concept achieved a
	higher relative TCAV score compared to every other concept, separately for
	the \texttt{Presence} and \texttt{Absence} classes.
	For this step, we used the same architecture, Pretrained ARN-50, to compute relative TCAV scores for both taxa in order to enable a direct comparison of concept rankings.

	\subsection{Robust TCAV results}
	We only show the scores for the \texttt{Presence} class, as they generally exhibit inverse behavior to those of the \texttt{Absence} class. Nonetheless, one must be aware that scores usually do not sum to 1. Full results for both the \texttt{Presence} and the \texttt{Absence} classes are available in the Appendix (Section~\ref{full_results}).
	
	\medskip
	\noindent\textbf{Robust TCAV.} The Robust TCAV scores of the 15 landscape concepts for Plecoptera and Trichoptera classification are displayed in Table~\ref{tab:combined_presence_tcav}. Following the Robust TCAV framework, scores above or below 0.5 indicate that a concept positively or negatively influences model predictions for a specific class, respectively. 
	
	\begin{table}[tb]
		\caption{Mean TCAV scores (with standard deviation) for the \texttt{Presence} class across all models and taxa.}
		\label{tab:combined_presence_tcav}
		\centering
		\small
		\begin{tabular}{l c@{\hskip 0.2cm}c@{\hskip 0.5cm} c@{\hskip 0.2cm}c@{\hskip 0.5cm} c@{\hskip 0.2cm}c}
			\toprule
			Concept & \multicolumn{2}{c}{CerberusCNN} & \multicolumn{2}{c}{ARN-50} & \multicolumn{2}{c}{PicoViT} \\
			\cmidrule(lr){2-3} \cmidrule(lr){4-5} \cmidrule(lr){6-7}
			& Pleco. & Trich. & Pleco. & Trich. & Pleco. & Trich. \\
			\midrule
			\texttt{Hedge}   & 0.30 (0.25) & 0.44 (0.14) & 0.90 (0.00) & 0.65 (0.00) & 0.85 (0.00) & 0.00 (0.00) \\
			\texttt{IsoTree} & 0.85 (0.02) & 0.74 (0.00) & 0.90 (0.00) & 0.65 (0.00) & 0.85 (0.00) & 0.00 (0.00) \\
			\texttt{Wood}    & 0.85 (0.00) & 0.74 (0.00) & 0.90 (0.00) & 0.65 (0.00) & 0.13 (0.06) & 0.01 (0.06) \\
			\texttt{LinW}    & 0.84 (0.08) & 0.40 (0.14) & 0.79 (0.28) & 0.65 (0.00) & 0.84 (0.04) & 0.00 (0.00) \\
			\texttt{SurfW}   & 0.15 (0.00) & 0.26 (0.00) & 0.10 (0.00) & 0.97 (0.09) & 0.16 (0.03) & 1.00 (0.00) \\
			\texttt{Wet}     & 0.85 (0.00) & 0.74 (0.00) & 0.90 (0.00) & 0.65 (0.00) & 0.85 (0.00) & 0.00 (0.00) \\
			\texttt{Road}    & 0.15 (0.00) & 0.79 (0.17) & 0.10 (0.00) & 0.65 (0.00) & 0.44 (0.08) & 0.00 (0.00) \\
			\texttt{Build}   & 0.15 (0.00) & 0.87 (0.05) & 0.10 (0.00) & 0.65 (0.00) & 0.15 (0.00) & 0.00 (0.00) \\
			\texttt{Wheat}   & 0.16 (0.04) & 0.26 (0.00) & 0.10 (0.00) & 0.35 (0.00) & 0.15 (0.01) & 0.00 (0.01) \\
			\texttt{Maize}   & 0.15 (0.00) & 0.32 (0.02) & 0.10 (0.00) & 0.35 (0.00) & 0.34 (0.06) & 0.80 (0.26) \\
			\texttt{Cereal}  & 0.15 (0.00) & 0.26 (0.00) & 0.10 (0.00) & 0.35 (0.00) & 0.25 (0.01) & 0.00 (0.00) \\
			\texttt{PermG}   & 0.85 (0.00) & 0.55 (0.08) & 0.90 (0.00) & 0.65 (0.00) & 0.85 (0.00) & 0.00 (0.00) \\
			\texttt{TempG}   & 0.85 (0.00) & 0.38 (0.08) & 0.30 (0.37) & 0.35 (0.00) & 0.85 (0.00) & 0.00 (0.00) \\
			\texttt{Organic} & 0.15 (0.02) & 0.23 (0.09) & 0.10 (0.00) & 0.35 (0.00) & 0.35 (0.11) & 0.00 (0.00) \\
			\texttt{Convent} & 0.15 (0.00) & 0.26 (0.00) & 0.10 (0.00) & 0.35 (0.00) & 0.36 (0.15) & 0.00 (0.00) \\
			\bottomrule
			\multicolumn{7}{l}{\scriptsize All scores are statistically significant ($p < 0.05$) against a $0.5$ null hypothesis via Student's t-test.}
		\end{tabular}
	\end{table}

	Across all architectures, \texttt{Presence} of Plecoptera is strongly associated with natural features (isolated tree, woodland, wetland, linear body of water, permanent grassland; over 0.85 TCAV scores), while anthropogenic features (building, road, crops) consistently negatively influence this class (below 0.16 TCAV scores). Trichoptera results vary by model. CerberusCNN and ARN-50 link \texttt{Presence} to natural features (isolated tree, woodland, wetland, permanent grassland; over 0.55 TCAV scores) but also to built surfaces (road and building). In contrast, these models interpret crops and temporary grassland as negative influence. Meanwhile, PicoViT predominantly yields scores of 0.00, except for surface water (1.00) and maize (0.80). For both taxa, organic and conventional crops, despite being opposite concepts, have similar TCAV scores across all models.
	
	With the exception of the results of PicoViT for Trichoptera, most \textbf{TCAV scores are consistent across architectures}: for both Plecoptera and Trichoptera, 12 concepts show concordant importance (all scores either above or below 0.5 across models). 
	Nevertheless, a few discrepancies emerge between architectures for specific concepts. For Plecoptera, hedgerow strongly positively influences the \texttt{Presence} class with ARN-50 (0.90) and PicoViT (0.85), yet it shows a negative correlation with CerberusCNN (0.30). Conversely, temporary grassland has a large TCAV score for the \texttt{Presence} of Plecoptera with CerberusCNN (0.85) while it has a small TCAV score with ARN-50 (0.30). 
	
	Also, several concepts influence both \texttt{Presence} and \texttt{Absence} simultaneously (scores over 0.5 for both). This occurs for road and building with CerberusCNN, and surface body of water with ARN-50 (0.97/0.95). Finally, while most standard deviations are near zero, significant instability is observed for hedgerow in CerberusCNN and temporary grassland and linear body of water with ARN-50.

	\begin{table}[tb]
		\caption{Concept importance ranking for Plecoptera and Trichoptera \texttt{Presence} classes.}
		\label{tab:ranking_presence}
		\centering
		\begin{tabular}{c@{\hskip 1cm}l@{\hskip 1cm}l}
			\toprule
			Rank & Plecoptera \texttt{Presence} & Trichoptera \texttt{Presence} \\
			\midrule
			1  & \texttt{Wood}    & \texttt{Wet}     \\
			2  & \texttt{Wet}     & \texttt{Wood}    \\
			3  & \texttt{PermG}   & \texttt{IsoTree} \\
			4  & \texttt{IsoTree} & \texttt{PermG}   \\
			5  & \texttt{Hedge}   & \texttt{Hedge}   \\
			6  & \texttt{LinW}    & \texttt{LinW}    \\
			7  & \texttt{TempG}   & \texttt{Build}   \\
			8  & \texttt{Wheat}   & \texttt{SurfW}   \\
			9  & \texttt{Organic} & \texttt{Road}    \\
			10 & \texttt{Build}   & \texttt{Organic} \\
			11 & \texttt{Road}    & \texttt{TempG}   \\
			12 & \texttt{SurfW}   & \texttt{Maize}   \\
			13 & \texttt{Convent} & \texttt{Convent} \\
			14 & \texttt{Maize}   & \texttt{Cereal}  \\
			15 & \texttt{Cereal}  & \texttt{Wheat}   \\
			\bottomrule
		\end{tabular}
	\end{table}
	
	\medskip
	\noindent\textbf{Relative importance.} Table \ref{tab:ranking_presence} provides a comparative ranking of concepts based on their relative influence across all models.
	For the \texttt{Presence} class, a high degree of convergence is observed between the two taxa: the top 6 concepts are nearly identical, with natural features such as woodland, wetland, permanent grassland, and isolated tree consistently driving classification. 
	The middle of the ranking diverges between Plecoptera and Trichoptera. For instance, building occupies a higher rank for Trichoptera \texttt{Presence} (Rank 7) compared to Plecoptera (Rank 10), and temporary grassland has a higher rank for Plecoptera \texttt{Presence} (Rank 7) compared to Trichoptera (Rank 11). We note that organic crops are consistently ranked higher than conventional crops. 
	A consistent inverse relationship is observed between the \texttt{Presence} and \texttt{Absence} rankings across both taxa. Features that occupy the top three positions for \texttt{Presence} (woodland, wetland, and permanent grassland) are positioned at the bottom of the \texttt{Absence} rankings. Conversely, crop concepts occupy the highest ranks for the \texttt{Absence} class while remaining among the lowest-ranked predictors for \texttt{Presence}.

	\section{Discussion}\label{discussion}
	As for the performance of SDMs, our study suggests that CNNs perform better than attention models, such as PicoViT, in predicting the \texttt{Presence} and \texttt{Absence} of Plecoptera and Trichoptera. While CerberusCNN and ARN-50 provided reliable results, PicoViT remained unreliable for Trichoptera. Local pixel neighborhoods likely contain more informative data regarding insect \texttt{Presence} than the long-distance relationships captured by Transformers \cite{dosovitskiy2021an,lecun2015deep}. Further experimentation is necessary to validate this finding. In addition, a context-adapted multiscale design allowed CerberusCNN to lead among models trained from scratch, though pretraining improved ARN-50, confirming the utility of deep CNNs even in data-limited contexts.
	
	Regarding the interpretation, \textbf{Robust TCAV produced highly encouraging results}. First, most scores are consistent across architectures, supporting the reliability of the results. Secondly, it suggests different effects depending on land use, aligning with the literature \cite{gomes2022does}. Natural concepts (\emph{e.g.}, woodland, wetland) appear to positively influence the prediction of Plecoptera and Trichoptera \texttt{Presence}. In contrast, crops (\emph{e.g.}, wheat, maize) consistently indicate \texttt{Absence}, also aligning with the literature on the danger of dominant agricultural practices (pesticides and fertilizers) for large sensitive taxa such as Plecoptera and Trichoptera~\cite{raitif2019stream}. Moreover, the greater positive influence of permanent over temporary grasslands reinforces the importance of natural elements for the \texttt{Presence} of insects. 
	
	A few non-trivial results also emerge. While both organic and conventional crops negatively influence insect \texttt{Presence} (scores < 0.5), organic practices consistently rank higher, supporting the expectation that they better sustain biodiversity \cite{hole2005does}. However, the negative influence of both may indicate that biodiversity-friendly practices at parcel-level may be undermined by degradation at landscape-level  \cite{allan2004landscapes}. The road and building concepts are linked to the \texttt{Presence} class of Trichoptera for both CerberusCNN and ARN-50. As the literature is not clear on this issue \cite{smith1983effect}, it might be interesting to further analyze it.
	
	Robust TCAV also generated inconsistent scores with respect to ecological processes. The inconsistent Robust TCAV scores obtained with PicoViT, which yielded near-zero values for both \texttt{Presence} and \texttt{Absence}, demonstrate that unreliable models provide meaningless ecological insights. Moreover, while Plecoptera and Trichoptera emerge from freshwater bodies \cite{ulfstrand1968life}, linear bodies of water and more significantly surface bodies of water are not recognized as primary influences for the \texttt{Presence} of both taxa. The difficulties of LiDAR with water surfaces and the canopy interference with multispectral sensors likely explain these results. These results are a good reminder that \textbf{concept-based XAI techniques only explain the model and not the ecological processes}. Finally, high standard deviation for TCAV scores for a few concepts (\emph{e.g.}, temporary grassland with respect to \texttt{Presence} of Plecoptera using ARN-50) suggests less reliable Concept Activation Vectors (CAVs) for these concepts.
	
	\section{Conclusion}\label{conclusion}
	\textbf{By leveraging our new landscape dataset, Robust TCAV successfully reunites the interpretation and prediction objectives of SDMs}, validating model logic against expert knowledge. This framework offers a new pathway to generate hypotheses in cases where results are counterintuitive (\emph{e.g.}, positive influence of buildings and roads on Trichoptera \texttt{Presence}). Finally, by providing clear, class-level explanations, our approach empowers policy-makers and land managers to make more informed, landscape-scale decisions. It is our hope that this landscape dataset will prove beneficial to the SDM community and that this work will serve as a catalyst for the adoption of concept-based XAI for SDMs. 
	
	Future work will explore additional concept-based XAI methods beyond Robust TCAV, such as Concept Activation Regions \cite{crabbe2022concept}, to compare how different frameworks capture ecological knowledge.
	
	\begin{credits}
		\subsubsection{\ackname}	This study was funded by the French National Research Agency (ANR), STRANGE project (ANR-23-CE03-0006). It has also benefited from the financial support from ANR project NIM-ML (ANR-21-CE23-0005-01) and from the international mobility grant program administered by the Doctoral College of Brittany and co-funded by the Brittany Region, as well as the logistical support of the Zone Atelier Loire, Zone Atelier Armorique, and Zone Atelier Brest-Iroise. We would like to also thank Thibaut Peres, Benjamin Bergerot and Laura Pellan for their precious contribution to the collection of field data.

	\subsubsection{\discintname}The authors have no competing interests to declare that are relevant to the content of this article.
	
	\subsubsection{AI Usage Declaration.}
	The author used LLMs to improve the readability and grammar of the manuscript, and enhance the clarity of the code (docstrings).
	
	\end{credits}
	
	\bibliographystyle{splncs04}
	\bibliography{biblio}
	\clearpage
	
	\appendix

	\section{Implementation resources}\label{ressources}

	All resources necessary to the implementation of our paper are listed below:
	\begin{itemize}
		\item \textbf{Concept dataset}: \url{https://zenodo.org/records/18936778}
		\item \textbf{Species distribution dataset}: \url{https://zenodo.org/records/18937048}
		\item \textbf{Sanity check dataset}: \url{https://zenodo.org/records/18939622}
		\item \textbf{Code}: \url{https://github.com/augustin-delabrosse/Robust-TCAV-for-SDM}
	\end{itemize}
	
	\section{Checking reliability of concept patches}\label{concept_sanity_check}
	To validate the distinctiveness of our concept representations, we first evaluate the separability of representative concepts against random patches. We select one concept from each major landscape category: \texttt{Hedge} (tree features), \texttt{Wet} (aquatic features), \texttt{Road} (infrastructure), \texttt{Wheat} (crops), and \texttt{TempG} (grasslands). Additionally, we directly compare \texttt{Organic} and \texttt{Convent} patches to assess whether the models can distinguish between visually similar farming practices with distinct ecological implications. We then perform a sanity check by applying Robust TCAV to verify that each concept shows a strong positive influence on its own classification, ensuring the internal consistency of our Concept Activation Vectors (CAVs).
	
	\subsection{Experimental design.}
	For each concept (with the exception of \texttt{Organic} and \texttt{Convent}), we extract from the multispectral data and the two LiDAR-derived elevation models 50 extra patches corresponding to the concept and 50 random patches that do not include the concept. Regarding \texttt{Convent} and \texttt{Organic}, we extract 50 patches for the former and 30 others for the latter. It is only possible to extract 30 for the latter as organic crops are under-represented in the landscape. After randomly splitting the patches between test (20\%), validation (16\%) and train (64\%) splits, we train a pretrained ARN-50 model (described in Section~\ref{models}) to classify \texttt{Presence} from \texttt{Absence} of the concept (binary classification). The training configuration is the same as the one described in Section~\ref{training_conf}. Finally, we use the test set to compute the Robust TCAV scores with respect to \texttt{Presence} and \texttt{Absence}. We expect scores to be high for \texttt{Presence} ($>$ 0.5) and low for \texttt{Absence} ($<$ 0.5).

	\subsection{Results.} When compared against random patches, most features achieve an AUC above 0.89  (see Table~\ref{tab:sanity_scores}). Temporary grasslands are moderately distinct from random patches with 0.65 AUC. The Robust TCAV scores are systematically over 0.5 with respect to the \texttt{Presence} of the concept. Furthermore, when \texttt{Hedge}, \texttt{Wet}, \texttt{Road} and \texttt{Wheat} are present, the scores are very high (> 0.88). Regarding \texttt{TempG} and \texttt{Organic}, scores related to \texttt{Presence} are moderately high (0.62 and 0.67, respectively). In contrast, the scores for the \texttt{Absence} class are always below 0.5. 
	
	\begin{table}[tb]
		\caption{Sanity Check: Discriminability (AUC) and Self-Influence (Robust TCAV) for selected landscape concepts.}
		\label{tab:sanity_scores}
		\centering
		\begin{tabular}{l@{\hskip 0.5cm}c@{\hskip 0.5cm}c@{\hskip 0.5cm}c}
			\toprule
			Concept & AUC & TCAV Presence & TCAV Absence \\
			\midrule
			\texttt{Hedge}        & 0.95 & 1.00 (0.00)    & 0.13 (0.00) \\
			\texttt{Road}         & 1.00 & 1.00 (0.00)    & 0.00 (0.00) \\
			\texttt{Wet}          & 0.89 & 0.88 (0.00)    & 0.25 (0.00) \\
			\texttt{TempG}        & 0.65 & 0.62 (0.00)    & 0.14 (0.00) \\
			\texttt{Wheat}        & 1.00 & 1.00 (0.00)    & 0.09 (0.00) \\
			\texttt{Organic}      & 0.92 & 0.67 (0.00)    & 0.00 (0.00) \\
			\texttt{Convent}      &      & 1.00 (0.00)    & 0.33 (0.00) \\
			\bottomrule
			\multicolumn{4}{l}{\scriptsize AUC is common to \texttt{Organic} and \texttt{Convent}}
		\end{tabular}
	\end{table}
	
	\subsection{Discussion.} \textbf{Our concepts demonstrate their high distinctiveness}. These concepts are simple objects (with the exception of \texttt{Organic} and \texttt{Convent}) that the multispectral data and the LiDAR easily manage to describe. This may explain the high AUC. The more moderate AUC for \texttt{TempG} (0.65) reflects the inherent ecological complexity of this concept. The concept of temporary grasslands gathers many different realities. Indeed, the age of temporary grasslands can vary from a few months (recently sown parcels) to 5 years (established meadows). This results in a highly heterogeneous signature.
	However, it should be noted that these results were obtained using a limited amount of training data. Moreover, the relative spatial proximity of patches may induce data leakage between the different splits. Therefore, they must be taken with caution.
	
	\textbf{The Robust TCAV scores confirm the internal consistency of our framework}. Indeed, most concepts are characterized by high scores with respect to their \texttt{Presence} and low scores with respect to their \texttt{Absence}, which indicates that the Concept Activation Vectors are robustly oriented toward the correct concept in the latent space. Even more subtle concepts, such as \texttt{TempG}, generate consistent TCAV scores, showing that the method is able to overcome intra-concept variability. Finally, the higher score obtained by \texttt{Convent} compared to \texttt{Organic} shows that the former concept is more stable than the latter. These results provide confidence that using our concepts with \textbf{Robust TCAV generates reliable explanations regarding the distribution of species}.

	\section{Distribution of training labels}\label{label_distrib}
	
	The distribution of labels for each taxon is given in Table~\ref{tab:presence_rates}.
	
	\begin{table}[tb]
		\caption{Presence rate for each taxon across the training, validation, and test splits.}
		\label{tab:presence_rates}
		\centering
		\begin{tabular}{lccc}
			\toprule
			Taxon & Train ($n=198$) & Validation ($n=38$) & Test ($n=48$) \\
			\midrule
			Plecoptera  & 53\% & 45\% & 42\% \\
			Trichoptera & 43\% & 45\% & 48\% \\
			\bottomrule
		\end{tabular}
	\end{table}
	
	\section{CerberusCNN architecture}\label{cerberuscnn_illustration}
	
\begin{figure}[tb]
		\includegraphics[width=\textwidth]{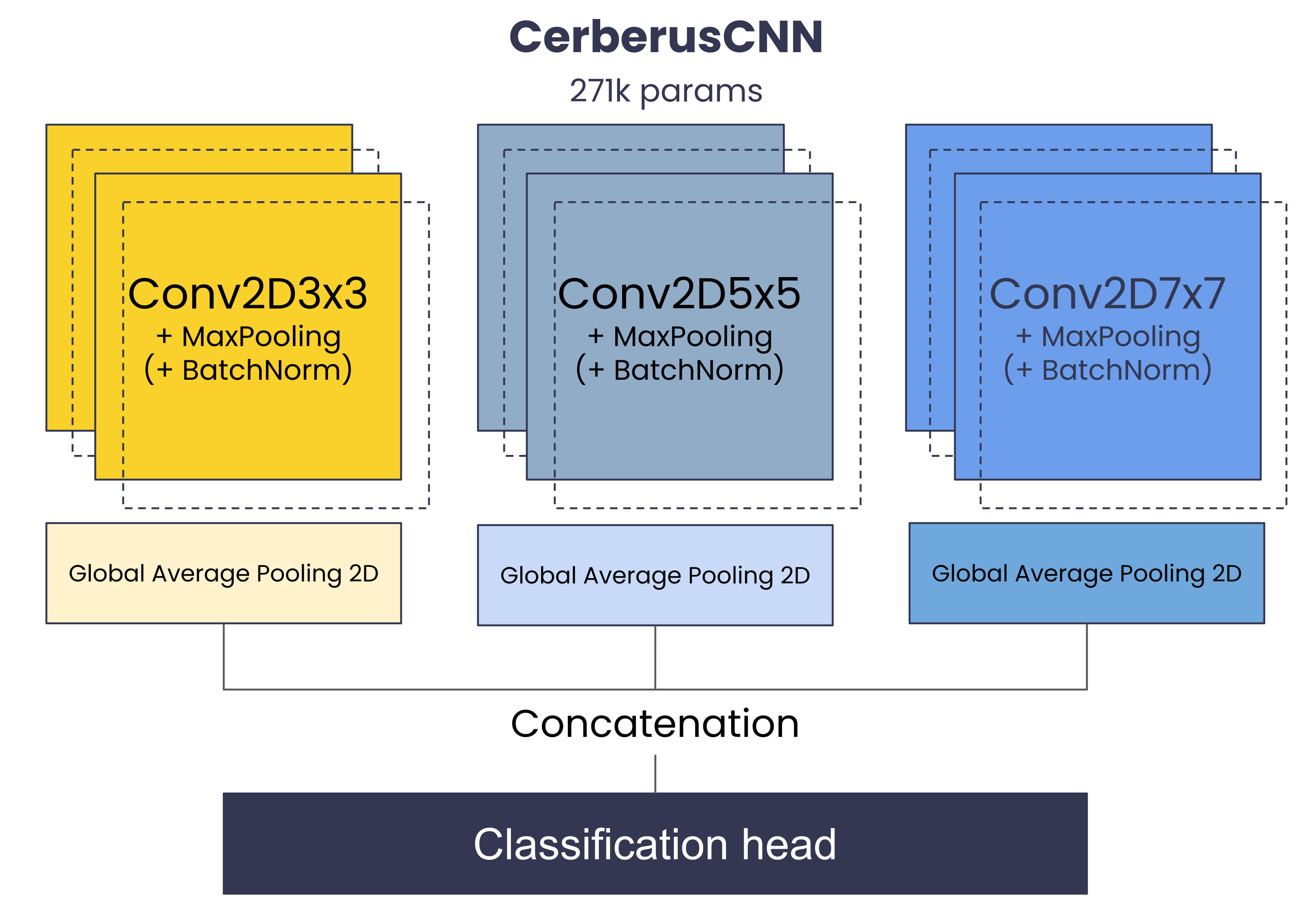}
		\caption{CerberusCNN is a convolutional architecture with three parallel branches to capture spatial features at multiple scales. The outputs of the branches are concatenated to feed a classification head.} \label{fig_cerberuscnn}
	\end{figure}

	CerberusCNN is a convolutional architecture with three parallel branches to capture spatial features at multiple scales, as illustrated in Fig~\ref{fig_cerberuscnn}. Each branch processes the input through two convolutional blocks with distinct kernel sizes (3, 5, and 7). The first convolutional block applies an initial max-pooling layer with distinct sizes (2, 4, and 8), while the second convolutional block applies a max-pooling layer of size 2 for all branches. The two blocks are separated by a batch normalization layer. Features are then aggregated via 2D global average pooling, concatenated, and processed through a classification head consisting of two dense layers (256 and 128 neurons) with intermediate dropout (0.5 and 0.3). It has approximately 270k learnable parameters.
	
	\section{Distribution maps}\label{classification_maps}
	
\begin{figure}[tb]
		\centering
		\includegraphics[width=0.75\textwidth]{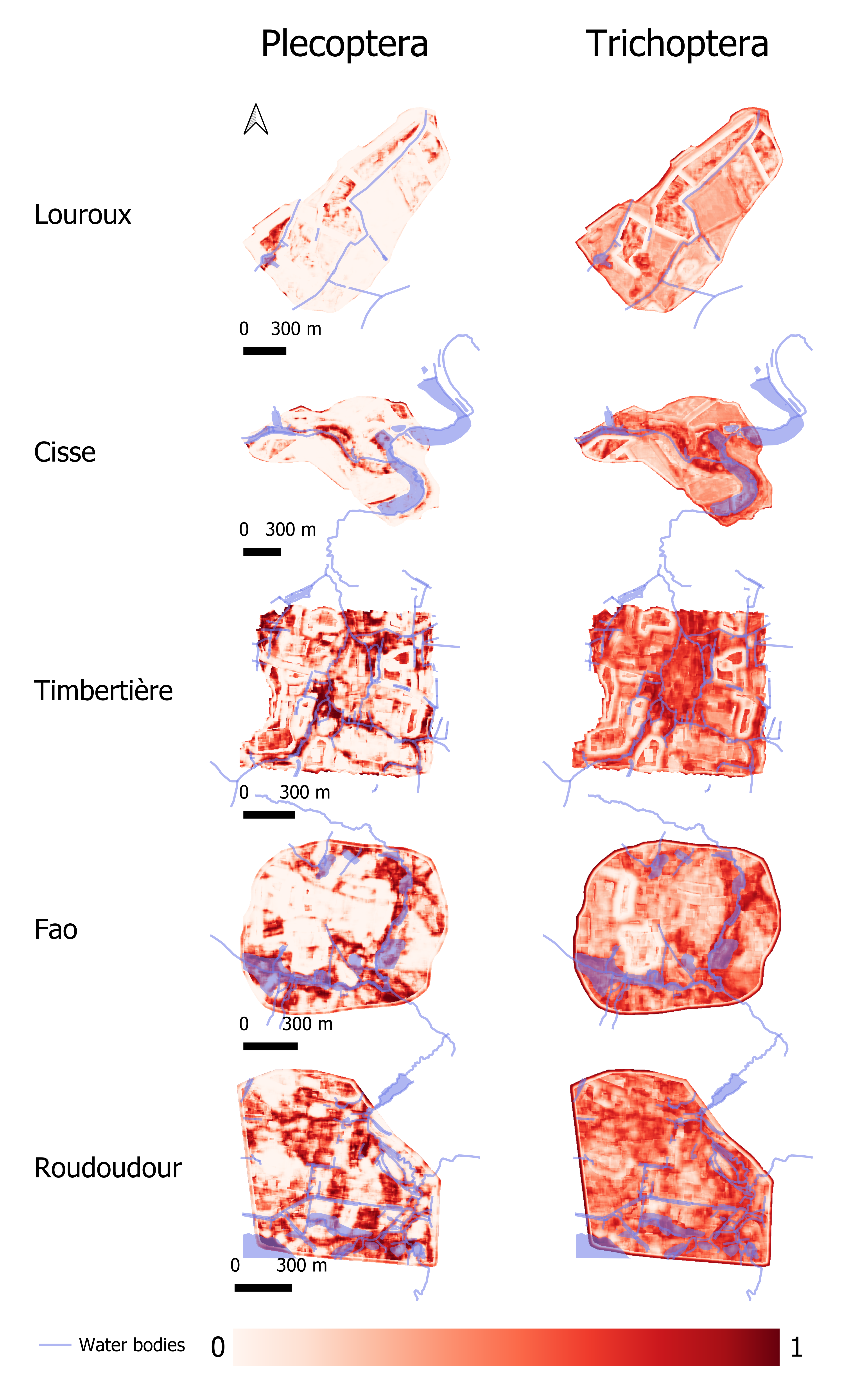}
		\caption{{Distribution maps} obtained with the CerberusCNN for the five study sites. The blue vectors represent the water bodies. The color scheme is as follows: dark red is used to denote high probability, and white is used to indicate low probability. Study site names are on the left.
		} \label{fig_distribution_maps}
	\end{figure}

	We created distribution maps (see Fig. \ref{fig_distribution_maps}) for Plecoptera and Trichoptera for each study site using CerberusCNN. To achieve this, we extract 512 $\times$ 512 patches and make a prediction for each pixel using a sliding window.
	
	\section{Full details for Robust TCAV and relative importance}\label{tutorial}
	
	\subsection{Step-by-step implementation.}
	
	The following protocol was used for each model to compute Robust TCAV scores:
	\begin{enumerate}
		\item \textbf{Select the layer.} We choose which layer $\ell$ will be used to extract the activation values and the gradient values. For CerberusCNN, we select the concatenation of the 3 branches; for ARN-50, we use the last bottleneck block before the average pooling layer; for PicoViT, we use the 2nd layer. The model can then be written as $f_{\ell} \circ h_{\ell, k}$, where $h_{\ell, k} : R^m \rightarrow R$ is the logit for class $k$ and $f_{\ell} : R^n \rightarrow R^m$ is the activation at layer $\ell$.
		\item \textbf{Gather gradients.} For each input patch $\mathbf{x}$, we compute the gradients at the selected layer $\ell$ with respect to the selected class $k$ logits $\nabla h_{\ell, k}(f_{\ell}(\mathbf{x}))$. These gradients are normalized by their $L^2$ norm.
		\item \textbf{Gather activations.} For both the selected concept $\mathbf{x}_C$ (\emph{e.g.}, hedgerows) and the random patches $\mathbf{x}_R$, we compute at layer $\ell$ their activations $f_{\ell}(\mathbf{x}_C)$ and $f_{\ell}(\mathbf{x}_R)$, respectively. 
		\item \textbf{Calculate mean activations.} We compute $\mu_C$ and $\mu_R$, respectively the mean of the concept activation vectors and of 500 randomly sampled random patches (50 could have been sufficient \cite{pmlr-v80-kim18d} but adding more increases the stability of the results \cite{wenkmann_garreau_2025}).
		\item \textbf{Build the Concept Activation Vector (CAV).} We obtain the CAV $v^{\ell}_C$ by subtracting the mean random activation vector from the mean concept activation vector, and then normalizing it by its $L^2$ norm.
		\item  \textbf{Compute sensitivity scores.} For each input patch, we obtain the sensitivity score:
		\begin{equation}
			S_{C,k,\ell}(\mathbf{x}) = \nabla h_{\ell, k}(f_{\ell}(\mathbf{x})) \cdot v^{\ell}_C.
		\end{equation} 
		Concept $C$ is considered to influence positively the prediction of class $k$ for input patch $x$ if $S_{C,k,\ell}(\mathbf{x}) > 0$. 
		\item \textbf{Calculate TCAV.} We assess the global importance of a concept for a given class (\emph{e.g.}, \texttt{Presence} or \texttt{Absence}) by computing the TCAV score. It is the proportion of input patches whose sensitivity scores are positive. A TCAV over 0.5 implies the model uses the concept to predict that class.
	\end{enumerate}
	
	We repeat 500 times steps 4 to 7 to get the mean and the standard deviation TCAV. 
	While some Robust TCAV implementations apply a non-zero threshold (typically 0.05) on the sensitivity score $S_{C,k,\ell}$ to eliminate low-magnitude signals, we found that a zero-threshold more accurately reflected the class-concept relationships in our specialized environmental models. More details are available below.
	
	\medskip
	\noindent After doing steps 1 and 2 described above, we follow these steps to obtain the relative importance of concepts:
	\begin{enumerate}
		\setcounter{enumi}{2}
		\item \textbf{Compute activations for all concepts.} We compute the activation at the selected layer for all concept patches using a forward hook.
		\item \textbf{Calculate mean activations per concept.} We compute the mean activation vectors for each concept.
		\item\textbf{Compare each pair of concepts $C_i$ and $C_j$:}
		\begin{enumerate}
			\item we subtract the mean activation vector of concept $C_j$ from that of concept $C_i$ to define direction $v^\ell_{{C_i},{C_j}}$ toward concept $C_i$. 
			\item For each input patch $\mathbf{x}$, we obtain a sensitivity score:
			\begin{equation}
				S_{C_i,C_j,k,\ell}(\mathbf{x}) = \nabla h_{\ell, k}(f_{\ell}(\mathbf{x})) \cdot v^\ell_{{C_i},{C_j}}.
			\end{equation} 
			\item We obtain the relative TCAV by computing the proportion of positive sensitivity scores for both \texttt{Presence} and \texttt{Absence} classes. A Relative TCAV over 0.5 for a given class indicates that the model leans towards the first concept $C_i$ (and conversely, if the relative TCAV is below 0.5).
		\end{enumerate}
		\item \textbf{Rank concepts.} For each concept, we count how many times it achieves a higher relative TCAV score compared to every other concept, separately for the \texttt{Presence} and \texttt{Absence} classes.
		
	\end{enumerate}
	
	\subsection{Results.}\label{full_results}
	
	Full results regarding Robust TCAV scores are displayed in Tables~\ref{tab:cerberuscnn_tcav}, \ref{tab:resnet_tcav} and \ref{tab:vit_tcav}. Full results regarding the relative importance of the concepts are given in Table~\ref{tab:ranking}.
	
	\begin{table}[tb]
		\caption{TCAV scores (mean with standard deviation in parentheses) for each concept and taxon for CerberusCNN.}
		\label{tab:cerberuscnn_tcav}
		\centering
		\begin{tabular}{l@{\hskip 0.5cm}c@{\hskip 0.5cm}c@{\hskip 0.5cm}c@{\hskip 0.5cm}c}
			\toprule
			Concept 
			& \multicolumn{2}{c@{\hskip 1cm}}{Plecoptera}
			& \multicolumn{2}{c}{Trichoptera} \\
			\cmidrule(lr){2-3} \cmidrule(lr){4-5}
			& Presence & Absence & Presence & Absence \\
			\midrule
			\texttt{Hedge}   & 0.30 (0.25) & 0.29 (0.13) & 0.44 (0.14) & 0.22 (0.04) \\
			\texttt{IsoTree} & 0.85 (0.02) & 0.21 (0.00) & 0.74 (0.00) & 0.28 (0.00) \\
			\texttt{Wood}    & 0.85 (0.00) & 0.21 (0.00) & 0.74 (0.00) & 0.28 (0.00) \\
			\texttt{LinW}    & 0.84 (0.08) & 0.21 (0.01) & 0.40 (0.14) & 0.20 (0.05) \\
			\texttt{SurfW}   & 0.15 (0.00) & 0.79 (0.00) & 0.26 (0.00) & 0.72 (0.00) \\
			\texttt{Wet}     & 0.85 (0.00) & 0.21 (0.00) & 0.74 (0.00) & 0.28 (0.00) \\
			\texttt{Road}    & 0.15 (0.00) & 0.79 (0.00) & 0.79 (0.17) & 0.80 (0.12) \\
			\texttt{Build}   & 0.15 (0.00) & 0.79 (0.00) & 0.87 (0.05) & 0.98 (0.03) \\
			\texttt{Wheat}   & 0.16 (0.04) & 0.79 (0.00) & 0.26 (0.00) & 0.72 (0.00) \\
			\texttt{Maize}   & 0.15 (0.00) & 0.79 (0.00) & 0.32 (0.02) & 0.14 (0.03) \\
			\texttt{Cereal}  & 0.15 (0.00) & 0.79 (0.00) & 0.26 (0.00) & 0.72 (0.00) \\
			\texttt{PermG}   & 0.85 (0.00) & 0.21 (0.00) & 0.55 (0.08) & 0.32 (0.08) \\
			\texttt{TempG}   & 0.85 (0.00) & 0.21 (0.00) & 0.38 (0.08) & 0.20 (0.01) \\
			\texttt{Organic} & 0.15 (0.02) & 0.76 (0.09) & 0.23 (0.09) & 0.12 (0.06) \\
			\texttt{Convent} & 0.15 (0.00) & 0.79 (0.00) & 0.26 (0.00) & 0.72 (0.00) \\
			\bottomrule
		\end{tabular}
	\end{table}
	
	\begin{table}[tb]
		\caption{TCAV scores (mean with standard deviation in parentheses) for each concept and taxon for ARN-50.}
		\label{tab:resnet_tcav}
		\centering
		\begin{tabular}{l@{\hskip 0.5cm}c@{\hskip 0.5cm}c@{\hskip 0.5cm}c@{\hskip 0.5cm}c}
			\toprule
			Concept 
			& \multicolumn{2}{c@{\hskip 1cm}}{Plecoptera}
			& \multicolumn{2}{c}{Trichoptera} \\
			\cmidrule(lr){2-3} \cmidrule(lr){4-5}
			& Presence & Absence & Presence & Absence \\
			\midrule
			\texttt{Hedge}   & 0.90 (0.00) & 0.25 (0.00) & 0.65 (0.00) & 0.28 (0.00) \\
			\texttt{IsoTree} & 0.90 (0.00) & 0.25 (0.00) & 0.65 (0.00) & 0.28 (0.00) \\
			\texttt{Wood}    & 0.90 (0.00) & 0.25 (0.00) & 0.65 (0.00) & 0.28 (0.00) \\
			\texttt{LinW}    & 0.79 (0.28) & 0.32 (0.18) & 0.65 (0.00) & 0.28 (0.00) \\
			\texttt{SurfW}   & 0.10 (0.00) & 0.75 (0.00) & 0.97 (0.09) & 0.95 (0.19) \\
			\texttt{Wet}     & 0.90 (0.00) & 0.25 (0.00) & 0.65 (0.00) & 0.28 (0.00) \\
			\texttt{Road}    & 0.10 (0.00) & 0.75 (0.00) & 0.65 (0.00) & 0.28 (0.00) \\
			\texttt{Build}   & 0.10 (0.00) & 0.75 (0.00) & 0.65 (0.00) & 0.28 (0.00) \\
			\texttt{Wheat}   & 0.10 (0.00) & 0.75 (0.00) & 0.35 (0.00) & 0.72 (0.00) \\
			\texttt{Maize}   & 0.10 (0.00) & 0.75 (0.00) & 0.35 (0.00) & 0.72 (0.00) \\
			\texttt{Cereal}  & 0.10 (0.00) & 0.75 (0.00) & 0.35 (0.00) & 0.72 (0.00) \\
			\texttt{PermG}   & 0.90 (0.00) & 0.25 (0.00) & 0.65 (0.00) & 0.28 (0.00) \\
			\texttt{TempG}   & 0.30 (0.37) & 0.73 (0.18) & 0.35 (0.00) & 0.72 (0.00) \\
			\texttt{Organic} & 0.10 (0.00) & 0.75 (0.00) & 0.35 (0.00) & 0.72 (0.00) \\
			\texttt{Convent} & 0.10 (0.00) & 0.75 (0.00) & 0.35 (0.00) & 0.72 (0.00) \\
			\bottomrule
		\end{tabular}
	\end{table}
	
	\begin{table}[tb]
		\caption{TCAV scores (mean with standard deviation in parentheses) for each concept and taxon for PicoViT.}
		\label{tab:vit_tcav}
		\centering
		\begin{tabular}{l@{\hskip 0.5cm}c@{\hskip 0.5cm}c@{\hskip 0.5cm}c@{\hskip 0.5cm}c}
			\toprule
			Concept 
			& \multicolumn{2}{c@{\hskip 1cm}}{Plecoptera}
			& \multicolumn{2}{c}{Trichoptera} \\
			\cmidrule(lr){2-3} \cmidrule(lr){4-5}
			& Presence & Absence & Presence & Absence \\
			\midrule
			\texttt{Hedge}   & 0.85 (0.00) & 0.33 (0.03) & 0.00 (0.00) & 0.00 (0.00) \\
			\texttt{IsoTree} & 0.85 (0.00) & 0.29 (0.00) & 0.00 (0.00) & 0.00 (0.00) \\
			\texttt{Wood}    & 0.13 (0.06) & 0.28 (0.05) & 0.01 (0.06) & 0.02 (0.07) \\
			\texttt{LinW}    & 0.84 (0.04) & 0.28 (0.01) & 0.00 (0.00) & 0.00 (0.00) \\
			\texttt{SurfW}   & 0.16 (0.03) & 0.71 (0.00) & 1.00 (0.00) & 1.00 (0.00) \\
			\texttt{Wet}     & 0.85 (0.00) & 0.29 (0.00) & 0.00 (0.00) & 0.00 (0.00) \\
			\texttt{Road}    & 0.44 (0.08) & 0.80 (0.02) & 0.00 (0.00) & 0.00 (0.00) \\
			\texttt{Build}   & 0.15 (0.00) & 0.71 (0.00) & 0.00 (0.00) & 0.00 (0.00) \\
			\texttt{Wheat}   & 0.15 (0.01) & 0.71 (0.00) & 0.00 (0.01) & 0.00 (0.01) \\
			\texttt{Maize}   & 0.34 (0.06) & 0.29 (0.04) & 0.80 (0.26) & 0.69 (0.29) \\
			\texttt{Cereal}  & 0.25 (0.01) & 0.71 (0.00) & 0.00 (0.00) & 0.00 (0.01) \\
			\texttt{PermG}   & 0.85 (0.00) & 0.29 (0.00) & 0.00 (0.00) & 0.00 (0.00) \\
			\texttt{TempG}   & 0.85 (0.00) & 0.29 (0.00) & 0.00 (0.00) & 0.00 (0.00) \\
			\texttt{Organic} & 0.35 (0.11) & 0.47 (0.06) & 0.00 (0.00) & 0.00 (0.00) \\
			\texttt{Convent} & 0.36 (0.15) & 0.76 (0.06) & 0.00 (0.00) & 0.00 (0.00) \\
			\bottomrule
		\end{tabular}
	\end{table}
	
	\begin{table}[tb]
		\caption{Concept ranking for Plecoptera and Trichoptera (\texttt{Presence} and \texttt{Absence} classes).}
		\centering
		\label{tab:ranking}
		\begin{tabular}{c@{\hskip 1cm}c@{\hskip 1cm}c@{\hskip 1cm}c@{\hskip 1cm}c}
			\toprule
			Rank & \multicolumn{2}{c@{\hskip 1cm}}{Plecoptera} & \multicolumn{2}{c}{Trichoptera} \\
			& Presence & Absence & Presence & Absence \\
			\cmidrule(lr){2-3} \cmidrule(lr){4-5}
			1  & \texttt{Wood}     & \texttt{Cereal}   & \texttt{Wet}      & \texttt{Wheat} \\
			2  & \texttt{Wet}      & \texttt{Maize}    & \texttt{Wood}     & \texttt{Cereal} \\
			3  & \texttt{PermG}    & \texttt{Convent}  & \texttt{IsoTree}  & \texttt{Convent} \\
			4  & \texttt{IsoTree}  & \texttt{SurfW}    & \texttt{PermG}    & \texttt{Maize} \\
			5  & \texttt{Hedge}    & \texttt{Organic}  & \texttt{Hedge}    & \texttt{TempG} \\
			6  & \texttt{LinW}     & \texttt{Road}     & \texttt{LinW}     & \texttt{Organic} \\
			7  & \texttt{TempG}    & \texttt{Build}    & \texttt{Build}    & \texttt{SurfW} \\
			8  & \texttt{Wheat}    & \texttt{Wheat}    & \texttt{SurfW}    & \texttt{Road} \\
			9  & \texttt{Organic}  & \texttt{TempG}    & \texttt{Road}     & \texttt{Build} \\
			10 & \texttt{Build}    & \texttt{LinW}     & \texttt{Organic}  & \texttt{PermG} \\
			11 & \texttt{Road}     & \texttt{Hedge}    & \texttt{TempG}    & \texttt{LinW} \\
			12 & \texttt{SurfW}    & \texttt{IsoTree}  & \texttt{Maize}    & \texttt{Hedge} \\
			13 & \texttt{Convent}  & \texttt{Wet}      & \texttt{Convent}  & \texttt{IsoTree} \\
			14 & \texttt{Maize}    & \texttt{PermG}    & \texttt{Cereal}   & \texttt{Wet} \\
			15 & \texttt{Cereal}   & \texttt{Wood}     & \texttt{Wheat}    & \texttt{Wood} \\
			\bottomrule
		\end{tabular}
	\end{table}
	
	\FloatBarrier
	\subsection{Analysis of the threshold.}\label{threshold}
	
	The Robust TCAV scores with a 0.05 threshold are given in Tables~\ref{tab:cerberuscnn_tcav_treshold}, \ref{tab:resnet_tcav_threshold} and \ref{tab:vit_tcav_threshold}.
	
	While Robust TCAV typically applies a threshold (e.g., 
	0.05) to eliminate low-magnitude signals \cite{martin2019interpretable}, our results indicate that a zero-threshold provides more ecologically relevant insights. To evaluate this, we compared the influence of all 15 concepts across our model-taxon combinations, totaling 75 instances (3 models for Plecoptera and 2 models for Trichoptera, excluding the PicoViT results for the latter) with and without a 0.05 threshold. Specifically, applying a 0.05 threshold resulted in 27 instances where a concept showed a similar influence (scores both > 0.5 or both < 0.5 with 0.5 being a neutral score) for both \texttt{Presence} and \texttt{Absence}, creating ambiguity. In contrast, using a zero-threshold reduced these ambiguous cases to only 12. In complex natural ecosystems, the distribution of aquatic insects often depends on a combination of subtle environmental features. By enforcing a threshold, these low-magnitude but biologically significant signals are prematurely discarded, obscuring the ability of the model to differentiate between occurrence classes. 
	
	\begin{table}[tb]
		\caption{TCAV scores (mean with standard deviation in parentheses) using a 0.05 threshold for each concept and taxon for CerberusCNN.}
		\label{tab:cerberuscnn_tcav_treshold}
		\centering
		\begin{tabular}{l@{\hskip 0.5cm}c@{\hskip 0.5cm}c@{\hskip 0.5cm}c@{\hskip 0.5cm}c}
			\toprule
			Concept 
			& \multicolumn{2}{c}{Plecoptera}
			& \multicolumn{2}{c}{Trichoptera} \\
			\cmidrule(lr){2-3} \cmidrule(lr){4-5}
			& Presence & Absence & Presence & Absence \\
			\midrule
			\texttt{Hedge}           & 0.12 (0.18) & 0.05 (0.05) & 0.31 (0.08) & 0.16 (0.04) \\
			\texttt{IsoTree}         & 0.85 (0.02) & 0.21 (0.00) & 0.74 (0.00) & 0.28 (0.00) \\
			\texttt{Wood}            & 0.85 (0.00) & 0.21 (0.00) & 0.74 (0.02) & 0.28 (0.00) \\
			\texttt{LinW}            & 0.81 (0.14) & 0.21 (0.03) & 0.28 (0.11) & 0.13 (0.06) \\
			\texttt{SurfW}           & 0.15 (0.00) & 0.78 (0.02) & 0.26 (0.00) & 0.72 (0.00) \\
			\texttt{Wet}             & 0.85 (0.00) & 0.21 (0.00) & 0.74 (0.00) & 0.28 (0.00) \\
			\texttt{Road}            & 0.15 (0.00) & 0.79 (0.00) & 0.56 (0.15) & 0.60 (0.11) \\
			\texttt{Build}           & 0.15 (0.00) & 0.79 (0.00) & 0.79 (0.03) & 0.85 (0.04) \\
			\texttt{Wheat}           & 0.15 (0.00) & 0.78 (0.00) & 0.26 (0.00) & 0.72 (0.00) \\
			\texttt{Maize}           & 0.15 (0.00) & 0.79 (0.00) & 0.22 (0.01) & 0.11 (0.02) \\
			\texttt{Cereal}          & 0.15 (0.00) & 0.79 (0.00) & 0.26 (0.00) & 0.72 (0.00) \\
			\texttt{PermG}           & 0.85 (0.00) & 0.21 (0.00) & 0.28 (0.06) & 0.18 (0.03) \\
			\texttt{TempG}           & 0.85 (0.00) & 0.21 (0.00) & 0.28 (0.04) & 0.17 (0.02) \\
			\texttt{Organic}         & 0.14 (0.02) & 0.71 (0.17) & 0.11 (0.08) & 0.03 (0.05) \\
			\texttt{Convent}         & 0.15 (0.00) & 0.79 (0.00) & 0.26 (0.00) & 0.72 (0.00) \\
			\bottomrule
		\end{tabular}
	\end{table}
	
	\begin{table}[tb]
		\caption{TCAV scores (mean with standard deviation in parentheses) using a 0.05 threshold for each concept and taxon for ARN-50.}
		\label{tab:resnet_tcav_threshold}
		\centering
		\begin{tabular}{l@{\hskip 0.5cm}c@{\hskip 0.5cm}c@{\hskip 0.5cm}c@{\hskip 0.5cm}c}
			\toprule
			Concept 
			& \multicolumn{2}{c}{Plecoptera}
			& \multicolumn{2}{c}{Trichoptera} \\
			\cmidrule(lr){2-3} \cmidrule(lr){4-5}
			& Presence & Absence & Presence & Absence \\
			\midrule
			\texttt{Hedge}           & 0.87 (0.17) & 0.24 (0.05) & 0.62 (0.14) & 0.27 (0.06) \\
			\texttt{IsoTree}         & 0.90 (0.00) & 0.25 (0.00) & 0.65 (0.00) & 0.28 (0.00) \\
			\texttt{Wood}            & 0.90 (0.00) & 0.25 (0.00) & 0.65 (0.00) & 0.28 (0.00) \\
			\texttt{LinW}            & 0.03 (0.17) & 0.01 (0.05) & 0.36 (0.32) & 0.15 (0.14) \\
			\texttt{SurfW}           & 0.02 (0.00) & 0.17 (0.00) & 0.00 (0.00) & 0.00 (0.00) \\
			\texttt{Wet}             & 0.90 (0.00) & 0.25 (0.00) & 0.65 (0.00) & 0.28 (0.00) \\
			\texttt{Road}            & 0.10 (0.00) & 0.75 (0.00) & 0.02 (0.12) & 0.01 (0.05) \\
			\texttt{Build}           & 0.10 (0.00) & 0.75 (0.00) & 0.28 (0.32) & 0.12 (0.14) \\
			\texttt{Wheat}           & 0.10 (0.00) & 0.75 (0.00) & 0.35 (0.00) & 0.72 (0.00) \\
			\texttt{Maize}           & 0.10 (0.00) & 0.75 (0.00) & 0.35 (0.00) & 0.72 (0.00) \\
			\texttt{Cereal}          & 0.10 (0.00) & 0.75 (0.00) & 0.35 (0.00) & 0.72 (0.00) \\
			\texttt{PermG}           & 0.90 (0.00) & 0.25 (0.00) & 0.64 (0.10) & 0.27 (0.04) \\
			\texttt{TempG}           & 0.00 (0.01) & 0.02 (0.11) & 0.35 (0.00) & 0.72 (0.00) \\
			\texttt{Organic}         & 0.10 (0.00) & 0.75 (0.00) & 0.02 (0.07) & 0.03 (0.15) \\
			\texttt{Convent}         & 0.10 (0.00) & 0.75 (0.00) & 0.35 (0.00) & 0.72 (0.00) \\
			\bottomrule
		\end{tabular}
	\end{table}
	
	\begin{table}[tb]
		\caption{TCAV scores (mean with standard deviation in parentheses) using a 0.05 threshold for each concept and taxon for PicoViT.}
		\label{tab:vit_tcav_threshold}
		\centering
		\begin{tabular}{l@{\hskip 0.5cm}c@{\hskip 0.5cm}c@{\hskip 0.5cm}c@{\hskip 0.5cm}c}
			\toprule
			Concept 
			& \multicolumn{2}{c}{Plecoptera}
			& \multicolumn{2}{c}{Trichoptera} \\
			\cmidrule(lr){2-3} \cmidrule(lr){4-5}
			& Presence & Absence & Presence & Absence \\
			\midrule
			\texttt{Hedge}           & 0.23 (0.08) & 0.02 (0.03) & 0.00 (0.00) & 0.00 (0.00) \\
			\texttt{IsoTree}         & 0.20 (0.04) & 0.04 (0.01) & 0.00 (0.00) & 0.00 (0.00) \\
			\texttt{Wood}            & 0.00 (0.00) & 0.00 (0.00) & 0.00 (0.00) & 0.00 (0.00) \\
			\texttt{LinW}            & 0.26 (0.10) & 0.09 (0.02) & 0.00 (0.00) & 0.00 (0.00) \\
			\texttt{SurfW}           & 0.10 (0.01) & 0.42 (0.06) & 0.00 (0.00) & 0.00 (0.00) \\
			\texttt{Wet}             & 0.45 (0.08) & 0.08 (0.04) & 0.00 (0.00) & 0.00 (0.00) \\
			\texttt{Road}            & 0.02 (0.03) & 0.11 (0.08) & 0.00 (0.00) & 0.00 (0.00) \\
			\texttt{Build}           & 0.15 (0.00) & 0.70 (0.02) & 0.00 (0.00) & 0.00 (0.00) \\
			\texttt{Wheat}           & 0.15 (0.00) & 0.55 (0.03) & 0.00 (0.00) & 0.00 (0.00) \\
			\texttt{Maize}           & 0.00 (0.01) & 0.00 (0.00) & 0.00 (0.00) & 0.00 (0.00) \\
			\texttt{Cereal}          & 0.12 (0.02) & 0.61 (0.03) & 0.00 (0.00) & 0.00 (0.00) \\
			\texttt{PermG}           & 0.80 (0.02) & 0.25 (0.01) & 0.00 (0.00) & 0.00 (0.00) \\
			\texttt{TempG}           & 0.81 (0.02) & 0.27 (0.02) & 0.00 (0.00) & 0.00 (0.00) \\
			\texttt{Organic}         & 0.01 (0.02) & 0.04 (0.03) & 0.00 (0.00) & 0.00 (0.00) \\
			\texttt{Convent}         & 0.11 (0.05) & 0.31 (0.15) & 0.00 (0.00) & 0.00 (0.00) \\
			\bottomrule
		\end{tabular}
	\end{table}

\end{document}